\documentclass{article}
\usepackage{ijcai19}

\usepackage{times}
\usepackage{soul}
\usepackage{url}
\usepackage[utf8]{inputenc}
\usepackage[small]{caption}
\usepackage{graphicx}
\usepackage{amsmath}
\usepackage{booktabs}
\usepackage{algorithm}
\usepackage{graphics}
\usepackage{graphicx}
\usepackage{epstopdf}
\usepackage{caption}
\usepackage{subcaption}
\usepackage{wrapfig}

\usepackage{enumitem}

\usepackage{tikz}
\usepackage{tabularx}
\usepackage{placeins}
\usepackage[hang,flushmargin]{footmisc}   

\usepackage{algorithm}
\usepackage{algpseudocode}
\usepackage{pifont}

\usepackage{amsmath}
\usepackage{amssymb}
\usepackage{amsfonts}
\usepackage{amsthm}
\usepackage{mathtools}
\usepackage{bm}
\usepackage{mathrsfs}

\usepackage{setspace}
\usepackage{xspace}

\newcommand{\Task}[1]{\mathcal{Z}^{(#1)}}

\newcommand{\Xt}[1]{\bm{X}^{(#1)}}

\newcommand{\Yt}[1]{\bm{Y}^{(#1)}}

\def\x{\mathbf{x}}

\def\x{{\mathbf x}}

\title{Complementary Learning for Overcoming Catastrophic Forgetting  Using Experience Replay}

\author{
Mohammad Rostami$^{1,2}$
\and
Soheil Kolouri$^2$\And
Praveen K. Pilly$^{2}$
\affiliations
$^1$University of Pennsylvania\\
$^2$HRL Laboratories, LLC\\
\emails
mrostami@seas.upenn.edu,
skolouri@hrl.com,
pkpilly@hrl.com
}

\begin{document}

\maketitle

\begin{abstract}
 Despite huge success, deep networks are unable to learn effectively in sequential multitask learning settings as they   forget the past learned tasks after learning new tasks. Inspired from complementary learning systems theory, we address this challenge by learning a generative model that couples the current task to the past learned tasks through a discriminative embedding space. We learn an abstract level generative distribution in the embedding which allows   generation of data points to represent past experience. We sample from this distribution and utilize experience replay to avoid forgetting and  simultaneously accumulate new knowledge to the abstract distribution in order to  couple the current task with past experience. We demonstrate theoretically and empirically that our framework learns a distribution in the embedding, which is shared across all task, and as a result tackles catastrophic
forgetting.

\end{abstract}

\section{Introduction}

Recent breakthrough of deep learning has led to algorithms with human-level performance  for many machine learning applications. This may seem natural as these networks are the best accessible tools that mimic human nervous system~\cite{morgenstern2014properties}. However, this success is highly limited to single task learning, and retaining learned knowledge in a continual learning setting remains a major challenge. That is, when a deep network is trained on multiple sequential tasks with diverse distributions, the new obtained knowledge usually interferes with past learned knowledge. As a result, the network often is unable to accumulate the new learned knowledge in a manner consistent with the past experience and forgets past learned tasks by the time the new task is learned. This phenomenon is called ``catastrophic forgetting" in the literature~\cite{french1999catastrophic}. This phenomenon is in contrast with continual learning ability of humans over their lifetime.

To mitigate catastrophic forgetting, a main  approach is to rely on replaying data points from past tasks that are stored selectively in a memory buffer~\cite{robins1995catastrophic}. This is inspired from the Complementary Learning Systems (CLS) theory~\cite{mcclelland1995there}. CLS theory hypothesizes that a dual long-term and short-term memory system, involving the neocortex and the hippocampus, is necessary for continual lifelong learning ability of humans.  In particular, the hippocampus rapidly encodes recent experiences as a short-term memory that is used to consolidate the knowledge in the slower neocortex as long-term memory through experience replays during sleep/conscious   recalls~\cite{diek2010}. Similarly,
if we selectively store samples from past tasks in a buffer, like in the neocortex, these samples can be replayed to the deep network in parallel with current task samples from recent-memory hippocampal storage to train the deep network jointly on past and current experiences. In other words, the online sequential learning problem is recast as an offline multitask learning problem that guarantees learning all tasks. A major issue with this approach is that the memory size grows as more tasks are learned and storing tasks' data points and updating the replaying process becomes more complex. Building upon recent successes of generative models~\cite{goodfellow2014generative}, this challenge has been addressed by amending the network structure such that it can generate  pseudo-data points for the past learned tasks without storing data points~\cite{shin2017continual}. 

In this paper, our goal is to address catastrophic forgetting via coupling sequential tasks in a latent  embedding space. We model this space as output space of a deep encoder, which is   between the input  and the output  layers of a deep classifier. Representations in this embedding space can be thought of neocortex  representations in the brain, which capture learned knowledge. To consolidate knowledge, we minimize the discrepancy between the distributions of all tasks in the embedding space.
 In order to mimic the offline memory replay process in the sleeping brain~\cite{rasch2013},
 we  amend the deep encoder with a decoder network to make the classifier network generative. The resulting autoencoding pathways
 can be thought of neocortical areas, which encodes and remembers past experiences. We fit a parametric distribution in the embedding space into the empirical distribution of data representations.  This distribution can be used to generate  pseudo-data points through sampling, followed by passing the samples into the decoder network. The pseudo-data points can then be used for experience replay of the previous tasks towards incorporation of new knowledge. This would enforce the embedding to be invariant with respect to the tasks as more tasks are learned, i.e., the network would retain the past learned knowledge as more tasks are learned.

\section{Related Work}

Past works have addressed catastrophic forgetting using two main approaches:   model consolidation~\cite{kirkpatrick2017overcoming} and experience replay~\cite{robins1995catastrophic}.
Both approaches implement a notion of memory to enable a network to remember the  distributions of   past learned tasks.

The idea of   model consolidation is based upon separating the information pathway for different tasks in the network such that new experiences do not interfere with past learned knowledge. This idea is inspired from the notion of structural plasticity~\cite{lamprecht2004structural}. During learning a task, important weight parameters for that task are identified and are consolidated when future tasks are learned. As a result, the new tasks are learned through free pathways in the network; i.e., the weights that are  important to retain knowledge about distributions of past tasks mostly remain unchanged. Several methods exist for identifying important weight parameters. Elastic Weight Consolidation (EWC) models posterior distribution of weights of a given network as a Gaussian distribution which is centered around the weight values from last  learned past tasks and a precision matrix, defined as the Fisher
information matrix of all network weights.
The weights then are consolidated   according to their importance, i.e., the value of Fisher coefficient~\cite{kirkpatrick2017overcoming}. In contrast to EWC, Zenke et al.~\cite{zenke2017continual} consolidate weights in an online scheme during learning a task. If a network weight contributes considerably to changes in the network loss, it is identified as an important weight. More recently, Aljundi et al.~\cite{aljundi2018memory} use a semi-Hebbian learning procedure to
  compute the importance of the weight parameters in a both unsupervised and online scheme.
The issue with the methods based on structural plasticity is that the network learning capacity is compromised 
to avoid catastrophic forgetting. As a result, the learning ability of the network decreases as more tasks are learned.

 Methods that use experience replay, use CLS theory  to retain the past tasks' distributions via replaying selected representative samples of past tasks continuously.  Prior works mostly have investigated on how to store a subset of past experiences to reduce dependence on memory.
 These samples can be selected in different ways. Schaul et al. select samples such that the effect of uncommon samples in the experience is maximized~\cite{schaul2015prioritized}. Isele and Cosgun explore four potential strategies to select more helpful samples in a buffer for replay~\cite{isele2018selective}. The downside is that storing samples requires memory and selection becomes more complex as more tasks are learned.
 To reduce dependence on a memory buffer similar to humans~\cite{french1999catastrophic}, Shin et al.~\cite{shin2017continual} developed a more efficient alternative by considering a generative model that can produce pseudo-data points of past tasks to avoid storing real data points. They use  a generative adversarial structures  to learn the tasks distributions to allow for generating pseudo-data points without storing data. However, adversarial learning is known to   require deliberate  architecture design  and selection of hyper-parameters~\cite{roth2017stabilizing}, and can suffer from  mode collapse~\cite{srivastava2017veegan}. Alternatively,
we demonstrate that  a simple  autoencoder structure can be used as the base generative model.  Our contribution is  to match the distributions of the tasks in the middle   layer of  the autoencoder and learn a shared distribution across the tasks to couple them. The shared distribution is then used to generate  samples for experience replay to avoid forgetting. We demonstrate effectiveness of our approach theoretically and empirically validate our method on benchmark tasks that has been used in the literature.

\section{Generative Continual Learning}

We consider a lifelong learning setting~\cite{chen2016lifelong}, where a learning agent  faces multiple, consecutive  tasks $\{\Task{t}\}_{t=1}^{T_{\text{Max}}}$ in a sequence $t=1, \ldots, T_{\text{Max}}$. The agent learns a new task at each time step and proceeds to learn the next task. Each task is learned based upon the  experiences, gained from learning past tasks. Additionally, the agent may encounter the learned tasks in future and hence must optimize its performance across all tasks, i.e., not to forget learned tasks when future tasks are learned. The agent also does not know a priori,  the total number of
tasks, which potentially might not be finite, distributions of the tasks, and the order of tasks.

\begin{figure}[t!]
    \centering
    \includegraphics[width=\linewidth]{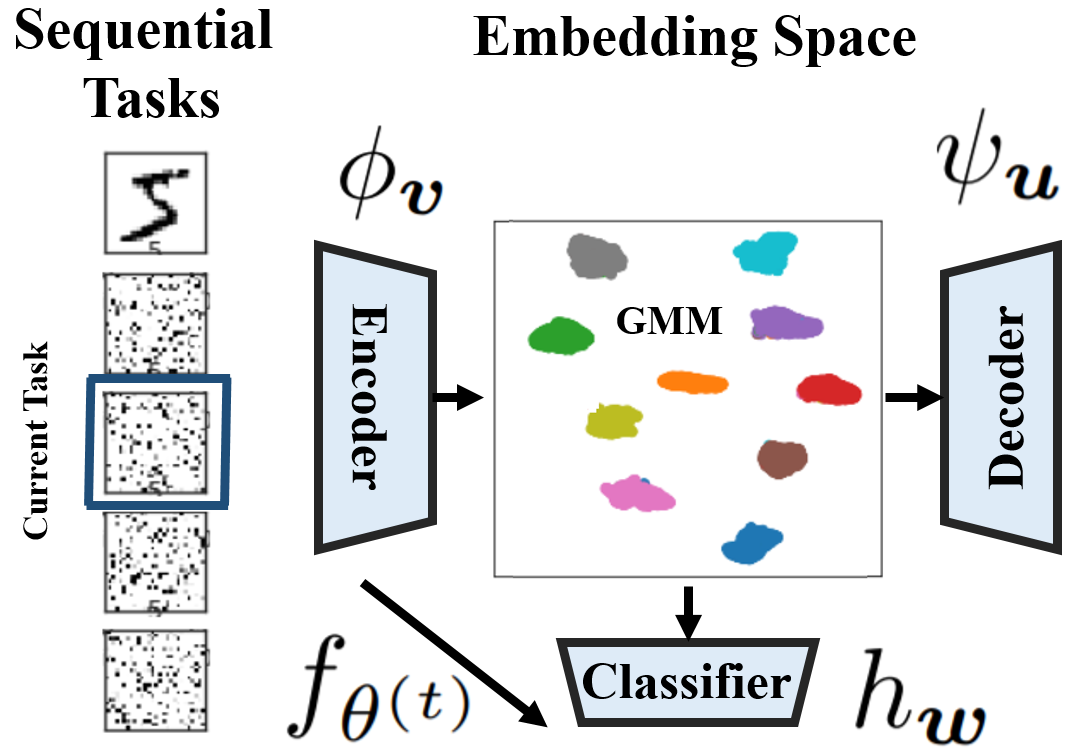}
         \caption{Architecture of the proposed   framework.}
         \label{structureFigCatfor}
\end{figure}

Let at time $t$, the current  task $\Task{t}$ with training dataset  $\Task{t} = \langle \Xt{t}, \Yt{t} \rangle$ arrives.  
We consider classification tasks where the training data points are drawn i.i.d. in pairs  from the   joint probability distribution, i.e., $(\bm{x}_i^{(t)},\bm{y}_i^{(t)})\sim p^{(t)}(\bm{x},\bm{y})$ which has the  marginal distribution $q^{(t)}$ over $\bm{x}$.  
 We assume that the lifelong learning agent trains a deep neural network $f_{\theta}:\mathbb{R}^{ d}\rightarrow \mathbb{R}^k$ with learnable weight parameters $\theta$ to map the data points $\Xt{t} = [\bm{x}_1^{(t)},\ldots, \bm{x}_{n_t}^{(t)}]\in\mathbb{R}^{d\times n_t}$ to the corresponding one-hot labels $\Yt{t}=[\bm{y}_1^{(t)},\ldots, \bm{y}_n^{(t)}]\in\mathbb{R}^{k \times n_t}$. 
 Learning a single  task in isolation is a standard classical learning problem. The agent can solve  for the optimal network weight parameters using standard empirical risk minimization (ERM), $\hat{ \theta}^{(t)}=\arg\min_{\theta}\hat{e}_{\theta}=\arg\min_{\theta}\sum_i \mathcal{L}_d(f_{\theta}(\bm{x}_i^{(t)}),\bm{y}_i^{(t)})$, where $\mathcal{L}_d(\cdot)$  is a proper loss function, e.g., cross entropy.
Given large enough number of labeled data points $n_t$, the model trained on a single task $\Task{t}$ will generalize well on the task test samples as  the empirical risk would be a suitable surrogate for the real risk function (Bayes optimal solution), $e = \mathbb{E}_{(\bm{x},\bm{y})\sim p^{(t)}(\bm{x},\bm{y})}(\mathcal{L}_d(f_{\theta^{(t)}}(\bm{x}),\bm{y}))$~\cite{shalev2014understanding}. The agent then can advance to learn the next task, but the challenge is that ERM is unable to tackle catastrophic forgetting as the model parameters are learned using solely the current task data which can potentially have very different distribution.    Catastrophic forgetting can be considered as the result of considerable deviations  of $\theta^{(T)}$ from past values over $\{\theta^{(t)}\}_{t=1}^{T-1}$ time as a result of  drifts in tasks' distributions $p^{(t)}(\bm{x},\bm{y})$. As a result, the updated $\theta^{(t)}$ can potentially be highly non-optimal for previous tasks. Our idea is  to prevent catastrophic forgetting through mapping all tasks' data into  an embedding space, where the tasks share a common distribution. We represent this space  by the output space of a deep network mid-layer  and we condition updating $\theta^{(t)}$ to what has been learned  before in this discriminative embedding space.  In other words, we want to train the deep network such the tasks are coupled in the embedding space through updating the parameter $\theta^{(T)}$ conditioned on $\{\theta^{(t)}\}_{t=1}^{T-1}$.

High performance of deep networks stems from learning data-driven and task-dependent high quality features~\cite{krizhevsky2012imagenet}. In other words,
a deep net  maps data points into a discriminative embedding space, captured by network layers, where classification can be performed easily, e.g., classes become separable in the embedding. Following this intuition, we   consider the deep net $f_{\theta}$ to be combined of  of  an encoder $\phi_{\bm{v}}(\cdot)$ with learnable parameter $\bm{v}$, i.e., early layers of the network,   and   a classifier network $h_{\bm{w}}(\cdot)$ with learnable parameters $\bm{w}$, i.e., higher layers of the network. The encoder sub-network $\phi_{\bm{v}}: \mathcal{X}\rightarrow \mathcal{Z}$ maps the data points into the embedding space $\mathcal{Z}\subset \mathbb{R}^f$ which describes the input in terms of abstract discriminative features. Note that after training,  as a deterministic function, the encoder network changes the input task data distribution. 

If the embedding space is discriminative, this distribution can be modeled as a multi-modal distribution for a given task, e.g., Gaussian mixture model (GMM). Catastrophic forgetting occurs because this distribution is not stationery with respect to different tasks.
The idea that we want to explore is based on training $\phi_{\bm{v}}$  such that all tasks share a similar distribution  in the embedding, i.e., the new tasks are learned such that their distribution in the embedding match the past experience, captured in the shared distribution.  Doing so, the embedding space becomes invariant with respect to any learned input task which in turn mitigates catastrophic forgetting.  

The key questions is how to adapt the standard supervised learning model  $f_{\theta}(\cdot)$ such that the embedding space, captured in the deep  network, becomes task-invariant. Following prior discussion, we use experience replay as the main strategy.  We expand the base network $f_{\theta}(\cdot)$  into a generative model by amending the model with a decoder $\psi_{\bm{u}}: \mathcal{Z}\rightarrow \mathcal{X}$, with learnable parameters $\bm{u}$. The encoder  maps the data representation back to the input space $\mathcal{X}$ and effectively make the pair $(\phi_{\bm{u}},\psi_{\bm{u}})$ an autoencoder. If implemented properly, we would learn a discriminative data distribution in the embedding space which can be approximated by a GMM. This distribution
captures our knowledge about past learned tasks.
When a new task arrives, pseudo-data points for past tasks can be generated by sampling from this distribution and feeding the samples to the decoder network. These pseudo-data points can be used for experience replay in order to tackle catastrophic forgetting. Additionally, we need to learn the new task such that its distribution matches the past shared distribution. As a result,  future pseudo-data points would represent the current task as well. Figure~\ref{structureFigCatfor} presents a  high-level block-diagram visualization of our framework.

\section{ Optimization Method}
 Following the above framework,  learning the first task ($t=1$) reduces     to minimizing  discrimination loss term for classification and reconstruction loss for the autoencoder to solve for optimal parameters $\hat{\bm{v}}^{(1)},\hat{\bm{w}^{(1)}}$ and $\hat{\bm{u}^{(1)}}$:
\footnotesize 
 \begin{equation}
\begin{split}
\min_{\bm{v},\bm{w},\bm{u}} \sum_{i=1}^{n_1} \mathcal{L}_d\Big(h_{\bm{w}}(\phi_{\bm{v}}\big(\bm{x}_i^{(1)})\big),\bm{y}_i^{(1)}\Big)+\gamma \mathcal{L}_r\Big(\psi_{\bm{u}}\big(\phi_{\bm{v}}(\bm{x}_i^{(1)})\big),\bm{x}_i^{(1)}\Big),
\end{split}
\label{eq:mainSupUnsupCatfor}
\end{equation}  
\normalsize
where  $\mathcal{L}_r$ is the reconstruction loss and $\gamma$ is a trade-off parameter between the two loss terms.

 Upon learning the first task, as well as subsequent future tasks, we can fit a GMM distribution with $k$ components to the  empirical distribution represented by data samples $\{(\phi_{\bm{v}}(\bm{x}_i^{(t)}),\bm{y}_i^{(t)})_{i=1}^{n_t}\}_{t=1}^T$ in the embedding space. The intuition behind this possibility is that as the embedding space is discriminative, we expect data points of each class form a cluster in the embedding.
 Let $\hat{p}_J^{(0)}(\bm{z})$ denote this parametric distribution. We update this distribution after learning each task to accumulative what has been learned from the new task to the distribution. As a result, this distribution captures knowledge about past. Upon learning this distribution, experience replay is feasible without saving data points. One can generate pseudo-data points in future through random sampling from $\hat{p}_J^{(T-1)}(\bm{z})$ at $t=T$ and then passing the samples through the decoder sub-network. 
 It is also crucial to learn the current task such that its    distribution in the embedding matches $\hat{p}_J^{(t-1)}(\bm{z})$. Doing so, ensures suitability of GMM to model the empirical distribution. The alternative approach is to use a Variational Auto encoder (VAE), but the discriminator loss helps forming the clusters automatically which in turn makes   normal autoencoders a feasible solution. 
 
 Let $\Task{T}_{ER} = \langle \Xt{T}_{ER}, \Yt{T}_{ER} \rangle$ 
 denote the pseudo-dataset, generated at $t = T$. Following our framework, learning subsequent tasks  reduces to solving the following   problem:
 \small
\begin{equation}
\begin{split}
&\min_{\bm{v},\bm{w},\bm{u}} \sum_{i=1}^{n_t} \mathcal{L}_d\Big(h_{\bm{w}}\big(\phi_{\bm{v}}(\bm{x}_i^{(t)})\big),\bm{y}_i^{(t)}\Big)+\gamma\mathcal{L}_r\Big(\psi_{\bm{u}}\big(\phi_{\bm{v}}(\bm{x}_i^{(t)})\big),\bm{x}_i^{(t)}\Big)\\&+\sum_{i=1}^{n_{er}} \mathcal{L}_d\Big(h_{\bm{w}}\big(\phi_{\bm{v}}(\bm{x}_{er,i}^{(T)})\big),\bm{y}_{er,i}^{(t)}\Big)+\gamma\mathcal{L}_r\Big(\psi_{\bm{u}}\big(\phi_{\bm{v}}(\bm{x}_{er,i}^{(t)})\big),\bm{x}_{er,i}^{(t)}\Big)\\&+\lambda \sum_{j=1}^k D\Big(\phi_{\bm{v}}(q^{(t)}(\bm{X}^{(t)}|C_j)),\hat{p}_J^{(t)}(\bm{Z}_{ER}^{(T)}|C_j)\Big),
\end{split}
\label{eq:mainPrMatchCatfor}
\end{equation}    
\normalsize
where $D(\cdot,\cdot)$ is a discrepancy measure, i.e., a metric, between two probability distributions and $\lambda$ is a trade-off parameter. The first four terms in Eq.~\eqref{eq:mainPrMatchCatfor} are  empirical  classification risk and autoencoder reconstruction loss terms for the current task and the generated pseudo-dataset. The third and the fourth term enforce learning the current task such that the past learned knowledge is not forgotten.  The fifth term is added to enforce the learned embedding distribution for the current task to be similar to what has been learned in the past, i.e., task-invariant. Note that we have  conditioned the distance between two distribution on classes to avoid class matching challenge, i.e., when wrong classes across two tasks are matched in the embedding, as well as to prevent mode collapse from happening. Class-conditional matching is feasible because we have labels for both distributions. Adding this term guarantees that we can continually use  GMM   to fit the shared distribution in the embedding space.

The main remaining question is selecting the metric $D(\cdot,\cdot)$  such that it fits our problem. Since we are computing the distance between empirical distributions, through drawn samples, we need a metric that can measure distances between distributions using the drawn  samples. Additionally, we must select a metric which has non-vanishing gradients as deep learning optimization  techniques are gradient-based methods. For these reasons, common distribution distance measures such as KL divergence and Jensen–Shannon divergence
are not suitable~\cite{kolouri2018sliced}. We rely on Wasserstein Distance (WD) metric~\cite{bonnotte2013unidimensional} which has been used extensively in deep learning applications. Since computing WD is computationally expensive, we use Sliced Wasserstein Distance (SWD)~\cite{rabin2011wasserstein} which   approximates WD, but can be computed efficiently.

SWD is computed through  slicing a high-dimensional  distributions. The $d$-dimensional distribution is decomposed into  one-dimensional marginal distributions by projecting the distribution into one-dimensional spaces that cover the high-dimensional  space. For a given distribution $p$,  a one-dimensional slice of the distribution is defined as:
\begin{equation}
\mathcal{R}p(t;\gamma)=\int_\mathcal{S} p(\bm{x})\delta(t-\langle\gamma, \bm{x}\rangle)d\bm{x},
\label{eq:radonCatfor}
\end{equation}
where $\delta(\cdot)$ denotes the Kronecker delta function,  $\langle \cdot ,\cdot\rangle$ denotes the vector dot product, $\mathbb{S}^{d-1}$ is the $d$-dimensional unit sphere and $\gamma$ is the projection direction. In other words, $\mathcal{R}p(\cdot;\gamma)$ is a marginal distribution of $p$  obtained from integrating $p$ over the hyperplanes orthogonal to $\gamma$.

SWD approximates the Wasserstein distance between two distributions    $p$ and $q$ by integrating the Wasserstein distances between the resulting sliced marginal distributions   of the two distributions over all $\gamma$:
\begin{eqnarray}
SW(p,q)=   \int_{\mathbb{S}^{d-1}} W(\mathcal{R} p(\cdot;\gamma),\mathcal{R} q(\cdot;\gamma))d\gamma
\label{eq:radonSWDdistanceCatfor}
\end{eqnarray}
 where $W(\cdot)$ denotes the Wasserstein distance.
 The main advantage of using SWD is that it can be computed efficiently as the  Wasserstein distance between one-dimensional  distributions has a closed form solution and  is equal to the $\ell_p$-distance between the inverse of their cumulative distribution functions. On the other hand, the $\ell_p$-distance between cumulative distribution can be approximated as the $\ell_p$-distance between the empirical cumulative distributions which makes SWD suitable in our framework. Finally, to approximate the integral in Eq.~\eqref{eq:radonSWDdistanceCatfor}, we relay on a Monte Carlo style integration   and approximate the SWD between $f$-dimensional samples $\{\phi_{\bm{v}}(\x_i^{(t)}\in \mathbb{R}^f\sim q^{(t)}\}_{i=1}^{n_t}$ and $\{\phi_{\bm{v}}(\x_{er,i}^{(t)})\in \mathbb{R}^f \sim \hat{p}_J^{(t)}\}_{j=1}^{n_t}$ in the embedding space as the following sum:
{\small
\begin{equation}
\begin{split}
&SW^2(\phi_{\bm{v}}(q^{(t)}),\hat{p}_J^{(t)}) \approx \\& \frac{1}{L}\sum_{l=1}^L \sum_{i=1}^{n_t}| \langle\gamma_l, \phi_{\bm{v}}(\x_{t_l[i]}^{(t)}\rangle)- \langle\gamma_l, \phi_{\bm{v}}(\x_{t_l[er,i]}^{(t)})\rangle|^2
\end{split}
\end{equation}}
where $\gamma_l\in\mathbb{S}^{f-1}$ denote  random samples that are drawn from the unit $f$-dimensional ball $\mathbb{S}^{f-1}$, and  $s_l[i]$ and $t_l[i]$ are the sorted indices of $\{\gamma_l\cdot\phi_{\bm{v}}(\x_i)\}_{i=1}^{n_t}$ for the two one-dimensional distributions. We utilize the SWD as the discrepancy measure between the    distributions in Eq.~\eqref{eq:mainPrMatchCatfor} to learn each task.  We tackle catastrophic forgetting using the proposed procedure. Our algorithm,   Continual Learning using Encoded Experience   Replay (CLEER) is summarized in Algorithm~\ref{GACLalgorithm}.

 \begin{algorithm}[tb!]
\caption{$\mathrm{CLEER}\left (L ,\lambda  \right)$\label{GACLalgorithm}} 
 {\small
\begin{algorithmic}[1]
\State \textbf{Input:} data $\mathcal{D}^{(t)}=(\bm{X}^{(t)},  \bm{Y}^{(t)})_{t=1}^{T_\text{Max}}$.
\State \textbf{Pre-training}: learning the first task ($t=1$)
\State \hspace{4mm} $\hat{ \theta}^{(1)}=(\bm{u}^{(1)},\bm{v}^{(1)},\bm{w}^{(1)}) =$$\arg\min_{\theta}\sum_i \mathcal{L}_d(f_{\theta}(\bm{x}_i^{(t)}),\bm{y}_i^{(t)})+\gamma \mathcal{L}_r(\psi_{\bm{u}}(\phi_{\bm{v}}(\bm{x}_i^{(1)})),\bm{x}_i^{(1)})$
\State Estimate $\hat{p}_J^{(0)}(\cdot)$ using $ \{\phi_{\bm{v}}(\bm{x}_i^{(1)}))\}_{i=1}^{n_t}$
\For{$t = 2,\ldots, T_{\text{Max}}$ }
\State \textbf{Generate pseudo dataset:}
\State  \hspace{4mm}  $\mathcal{D}_{\text{ER}} = 
\{(\bm{x}_{er,i}^{(t)}=\psi(\bm{z}_{er,i}^{(t)}), \bm{y}_{er,i}^{(t)})\sim \hat{p}_J^{(t-1)}(\cdot) \}_{i=1}^{n_{er}}$
\State \textbf{Update} learnable parameters using pseudo dataset:) Eq.~\eqref{eq:mainPrMatchCatfor}
\State \textbf{Estimate:} $\hat{p}_J^{(t)}(\cdot)$
\State \hspace{4mm} use $ \{\phi_{\bm{v}}(\bm{x}_i^{(t)})),\phi_{\bm{v}}(\bm{x}_{er,i}^{(t)}))\}_{i=1}^{n_t}$
\EndFor
\end{algorithmic}}
\end{algorithm}

\section{Theoretical Justification}
We use existing theoretical results about using optimal transport within domain adaptation~\cite{redko2017theoretical}, to justify why our algorithm can tackle catastrophic forgetting. Note that the hypothesis class in our learning  problem  is the set of all functions represented by the network  $f_{\theta}(\cdot)$ parameterized by   $\theta$. For a given model in this  class, let $e_{t}$ denote the observed risk for a particular task $\Task{t}$ and $e^J_{t}$ denote the observed risk for learning the network on samples of the distribution $\hat{p}_J^{(t-1)}$. We rely on the following theorem.

\textbf{Theorem~1 \cite{redko2017theoretical}}: Consider two tasks $\Task{t}$ and $\Task{t'}$, and a model $f_{\theta^{(t')}}$  trained for $\Task{t'}$, then for any $d'>d$ and $\zeta<\sqrt{2}$, there exists a constant number $N_0$ depending on $d'$ such that for any  $\xi>0$ and $\min(n_t,n_{t'})\ge \max (\xi^{-(d'+2),1})$ with probability at least $1-\xi$ for all $f_{\theta^{(t')}}$, the following holds:
\begin{equation}
\begin{split}
e_{t}\le & e_{t'} +W(\hat{p}^{(t)}, \hat{p}^{(t')})+e_{\mathcal{C}}(\theta^*)+\\
& \sqrt{\big(2\log(\frac{1}{\xi})/\zeta\big)}\big(\sqrt{\frac{1}{n_t}}+\sqrt{\frac{1}{n_{t'}}}\big),
\end{split}
\label{eq:theroemfromcourtyCatFor}
\end{equation}    
where $W(\cdot)$ denotes the Wasserstein distance between empirical distributions of the two tasks and $\theta^*$ denotes the optimal parameter for  training the model on tasks jointly, i.e., $\theta^*= \arg\min_{\theta} e_{\mathcal{C}}(\theta)=\arg\min_{\theta}\{ e_{t}+  e_{t'}\}$. 

We observe from Theorem~1 that performance, i.e., real risk, of a model learned for task $\Task{t'}$ on another  task $\Task{t}$ is upper-bounded by four terms: i) model performance on task $\Task{t'}$, ii) the distance between the two distributions, iii) performance of the jointly learned model $f_{\theta^*}$, and iv) a constant term which depends on the number of data points for each task.
Note that we do not have a notion of time in this Theorem and the roles of $\Task{t}$ and $\Task{t'}$ 
can be shuffled and the theorem would still hold. In our framework, we consider the task $\Task{t'}$, to be the pseudo-task, i.e., the task derived by drawing samples from $\hat{p}_J^{t'}$ and then feeding  the samples to the decoder sub-network.  We use this result to conclude the following lemma.

\textbf{Lemma~1 }: Consider CLEER algorithm for lifelong learning after  $\Task{T}$ is learned at time $t=T$. Then all tasks $t<T$ and under the conditions of Theorem~1, we can conclude the  following inequality:
\begin{equation}
\begin{split}
e_{t}\le & e_{T-1}^J +W(\hat{q}^{(t)}, \psi(\hat{p}_J^{(t)}))+\sum_{s=t}^{T-2} W(\psi(\hat{p}_J^{(s)}), \psi(\hat{p}_J^{(s+1)}))\\
& +e_{\mathcal{C}}(\theta^*)+\sqrt{\big(2\log(\frac{1}{\xi})/\zeta\big)}\big(\sqrt{\frac{1}{n_t}}+\sqrt{\frac{1}{n_{er,t-1}}}\big),
\end{split}
\label{eq:theroemfromcourtyCatForours}
\end{equation}    
Proof: We consider $\Task{t}$ with empirical distribution $\hat{q}^{(t)}$  and the pseudo-task with the distribution  $\psi(\hat{p}_J^{(T-1)})$ in the network input space, in Theorem~1. Using the triangular inequality on the term $W(\hat{q}^{(t)}, \psi(\hat{p}_J^{(T-1)}))$ recursively, i.e., $W(\hat{q}^{(t)}, \psi(\hat{p}_J^{(s)}))\le W(\hat{p}^{(t)}, \psi(\hat{p}_J^{(s-1)}))+W(\psi(\hat{p}_J^{(s)}), \psi(\hat{p}_J^{(s-1)}))$ for all $t \le s< T$,  Lemma~1 can be derived.

Lemma~1 explains why our algorithm can tackle catastrophic forgetting. When future tasks are learned, our algorithms updates the model parameters  conditioned on minimizing  the upper bound of $e_{t}$ in Eq.~\ref{eq:theroemfromcourtyCatForours}. Given suitable network structure and in the presence of enough labeled data points, the terms  $e_{t-1}^J$ and $e_{\mathcal{C}}(\theta^*)$ are minimized using ERM, and the last constant term would be small. The term $W(\hat{q}^{(t)}, \psi(\hat{p}_J^{(t)}))$ is minimal because we deliberately fit  
the distribution $\hat{p}_J^{(t)}$ to the distribution  $\phi(\hat{q}^{(t)})$ in the embedding space and ideally learn $\phi$ and $\psi$ such that $\psi \approx \phi^{-1}$. This term demonstrates that minimizing the discrimination loss is critical as only then, we can fit a GMM distribution on $\phi(\hat{p}^{(t)})$   with high accuracy.  Similarly, the sum terms in Eq~\ref{eq:theroemfromcourtyCatForours} are minimized because at $t=s$ we draw samples from $\hat{p}_J^{(s-1)}$ and enforce indirectly  $\hat{p}_J^{(s-1)} \approx \phi(\psi(\hat{p}_J^{(s-1)}))$. Since the upper bound of $e_{t}$  in Eq~\ref{eq:theroemfromcourtyCatForours} is minimized and  conditioned on tightness of the upper bound, the task $\Task{t}$ will not be forgotten.

\section{Experimental Validation}

 We validate  our method on learning two sets of sequential tasks: independent permuted MNIST tasks and related digit classification tasks.  

\subsection{Learning sequential independent tasks}

 Following the literature, we use 
permuted MNIST tasks to validate our framework. The sequential tasks involve classification of handwritten images of MNIST ($\mathcal{M}$) dataset~\cite{lecun1990handwritten}, where pixel values for each data point are shuffled randomly by a fixed   permutation order for each task.
As a result, the tasks are  independent and quite different from each other. Since knowledge transfer across tasks is less likely to happen, these tasks are a suitable benchmark to investigate the effect of an algorithm on mitigating catastrophic forgetting as past learned tasks are not  similar to the current task. We  compare our method against: a) normal back propagation (BP) as a lower bound, b) full experience replay (FR) of data for all the previous tasks  as an upper bound, and c) EWC as a competing weight consolidation framework.  
\begin{figure}[tb!]
    \centering
    \vspace{-10pt}
    \hspace{-10pt}
           \begin{subfigure}[b]{0.22\textwidth}\includegraphics[width =\textwidth,]{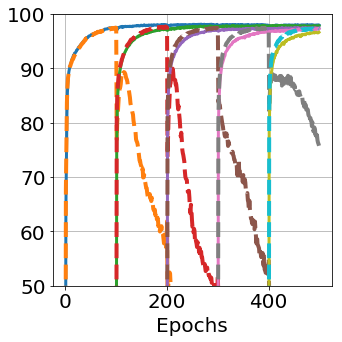}
           \vspace{-6mm}
           \centering
        \caption{BP vs EWC}
        \label{fig:Catfor_zero}
    \end{subfigure}
    \begin{subfigure}[b]{0.22\textwidth}\includegraphics[width=\textwidth]{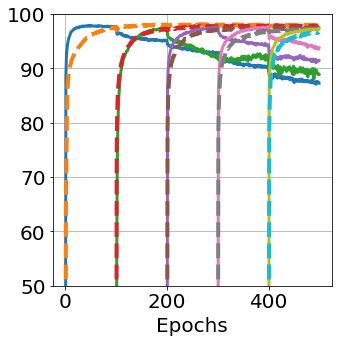}
           \centering
           \vspace{-6mm}
        \caption{CLEER vs FR}
        \label{fig:Catfor_Full}
    \end{subfigure}
    \vspace{-4mm}
     \caption{Performance results and  for permuted MNIST   tasks.  (Best viewed in color.) }\label{fig:resultsCatforgetPMNIST}
\end{figure}

\begin{figure}[tb!]
    \centering
    \vspace{-10pt}
    \hspace{-10pt}
    \begin{subfigure}[b]{0.22\textwidth}\includegraphics[width=\textwidth]{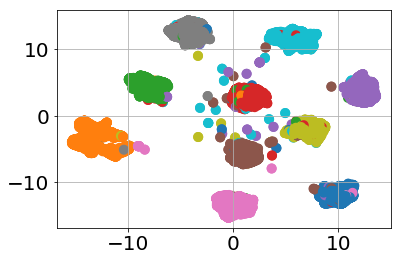}
           \centering
           \vspace{-6mm}
        \caption{ CLEER}
        \label{fig:Catfor_EWC}
    \end{subfigure}
       \begin{subfigure}[b]{0.22\textwidth}\includegraphics[width=\textwidth]{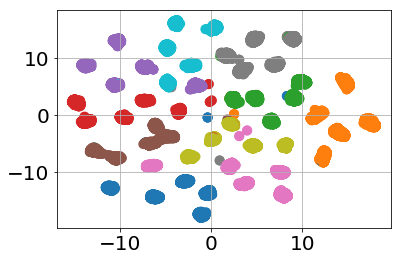}
              \centering
           \vspace{-6mm}
        \caption{  FR}
        \label{fig:Catfor_Ours}
    \end{subfigure}
    \vspace{-4mm}
     \caption{UMAP visualization of CLEER versus FR  for MNIST permutation tasks.  (Best viewed in color.) }\label{fig:resultsCatforgetPMNISTembed}
\end{figure}

  We learn permuted MNIST  using a simple multi layer perceptron (MLP) network trained via standard stochastic gradient descent  and at each iteration, compute the performance of the network on the testing split of each task data. 
Figure~\ref{fig:resultsCatforgetPMNIST} presents  results on five permuted MNIST tasks. Figure~\ref{fig:Catfor_zero} presents learning curves for BP (dotted curves) and EWC (solid curves)~\footnote{We have used  PyTorch  implementation of EWC~\cite{WinNT}. }. We observe that EWC is able to address catastrophic forgetting quite well. But a  close inspection reveals that as more tasks are learned, the asymptotic performance on  subsequent tasks is  less than the single task learning performance (roughly $4 \%$ less for the fifth task). This can be understood as a side  effect of weight consolidation which limits the learning capacity of the network. This in an inherent limitation for techniques that regularize network parameters to prevent catastrophic forgetting. Figure~\ref{fig:Catfor_Full} presents learning curves for our method (solid curves) versus FR (dotted curves). As expected, FR can prevent catastrophic forgetting perfectly but as we discussed the downside is memory growth challenge.  FR result in Figure~\ref{fig:Catfor_Full} demonstrates that the network learning capacity is sufficient for learning these tasks and if we have a perfect generative model, we can prevent catastrophic forgetting without compromising the network learning capacity. Despite more forgetting in our approach compared to EWC, the asymptotic performance after learning each task, just before advancing to learn the next task, has been improved. 
We also observe that our algorithm suffers an initial drop in performance of previous tasks, when we proceed to learn a new task. Forgetting beyond this initial forgetting is negligible. This can be understood as  the existing distance between $\hat{p}_J^{(T-1)}$ and $\phi(q^{(t)})$ at $t=T$. In other words, our method can be improved if more advanced autoencoder structures are used.
These results may suggest that catastrophic forgetting may be tackled better if both weight consolidation and experience replay are combined.

\begin{figure}[tb!]
    \centering
    \vspace{-10pt}
    \hspace{-10pt}
           \begin{subfigure}[b]{0.22\textwidth}\includegraphics[width =\textwidth,]{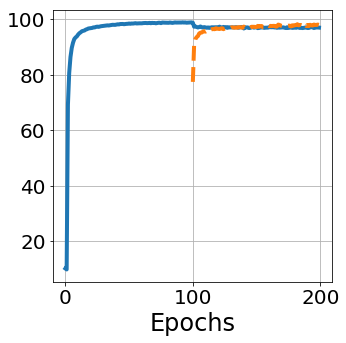}
           \centering
        \caption{$\mathcal{M}\rightarrow \mathcal{U} $}
        \label{fig:USPSSVHNCatfor_zero}
    \end{subfigure}
    \begin{subfigure}[b]{0.22\textwidth}\includegraphics[width=\textwidth]{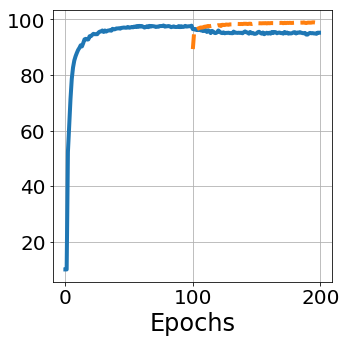}
           \centering
        \caption{$\mathcal{U}\rightarrow \mathcal{M} $}
        \label{fig:USPSSVHNCatfor_Full}
    \end{subfigure}
     \caption{Performance results on MNIST and USPS digit recognition tasks.  (Best viewed in color.)}\label{fig:resultsCatforgetRelated}
\end{figure}

\begin{figure}[tb!]
    \centering
    \vspace{-10pt}
    \hspace{-10pt}
    \begin{subfigure}[b]{0.22\textwidth}\includegraphics[width=\textwidth]{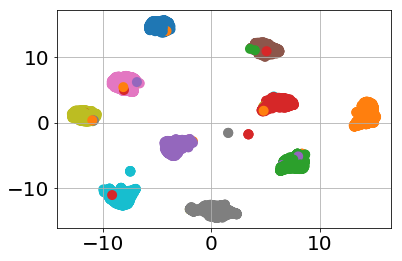}
           \centering
        \caption{$\mathcal{M}\rightarrow \mathcal{U} $}
        \label{fig:USPSSVHNCatfor_EWC}
    \end{subfigure}
       \begin{subfigure}[b]{0.22\textwidth}\includegraphics[width=\textwidth]{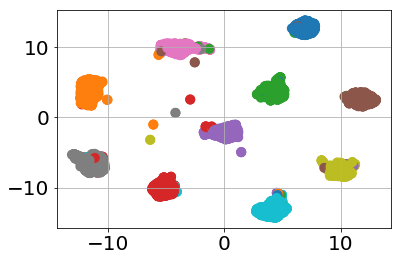}
              \centering
        \caption{$\mathcal{U}\rightarrow \mathcal{M} $}
        \label{fig:USPSSVHNCatfor_Ours}
    \end{subfigure}
     \caption{UMAP visualization for $\mathcal{M}\rightarrow \mathcal{U} $ and $\mathcal{U}\rightarrow \mathcal{M} $ tasks.  (Best viewed in color.)}\label{fig:resultsCatforgetRelatedembed}
\end{figure}

To provide a better intuitive understating,
we have also included the representations of the testing data for all tasks in the embedding space of the neural network in Figures~ \ref{fig:resultsCatforgetPMNISTembed}. We have used UMAP~\cite{mcinnes2018umap} to reduce the dimension for visualization purpose.
 In these figures, each color corresponds to a specific class of digits. We can see that although FR is able to learn all tasks  and form distinct clusters for each digit for each task, but five different clusters are formed for each class in the embedding space. This suggests that FR is unable to learn the concept of the same class across different tasks in the embedding space. In comparison, we observe that CLEER is able to match the same class across different tasks, i.e., we have exactly ten clusters for the ten digits. This empirical observation demonstrates that we can model the data distribution in the embedding using 
a multi-modal distribution such as GMM~\cite{heinen2012using}.

 \subsection{Learning  sequential  tasks in related domains}
We performed a second set of experiments on related tasks to investigate the ability of the algorithm to learn new domains. We consider two  digit classification datasets for this purpose: MNIST ($\mathcal{M}$) and
USPS ($\mathcal{U}$) datasets. Despite being  similar,  USPS dataset is a more challenging task as the number of training set is smaller, 20,000 compared to 60,000 images. We   consider the two possible sequential learning scenarios $\mathcal{M}\rightarrow \mathcal{U} $ and $\mathcal{M}\rightarrow \mathcal{U} $.    We  resized the USPS images    to $28\times 28$ pixels   to be able to use  the same encoder network for both tasks.   The experiments can be considered as a special case of  domain adaptation as both tasks are digit recognition tasks but in different domains. To capture relations between the tasks, we use a convolutional network for these experiments.

Figure~\ref{fig:resultsCatforgetRelated} presents learning curves for these two tasks. We observe that the network retains the knowledge about the first domain, after learning the second domain. We  observe  that forgetting is negligible compared to unrelated tasks and we have jump-start performance boost. These observations suggests relations between the tasks helps to avoid forgetting.  As a result of task similarities, the empirical distribution $\hat{p}_J^{(1)}$ can capture the task distribution more accurately. As expected from the theoretical justification, this empirical result suggests the performance of our algorithm depends on closeness of the distribution $\psi(\hat{p}_J^{(t)})$ to the distributions of previous tasks and improving probability estimation will increase performance of our approach.  
We have also presented UMAP visualization of  all tasks data in the embedding space   in Figures~ \ref{fig:resultsCatforgetRelatedembed}.  We can see that as expected the distributions are matched in the embedding space.

\section{Conclusions}

Inspired from CLS theory, we addressed the challenge of catastrophic forgetting for  sequential learning of multiple tasks using experience replay. We amend a base learning model with  a generative pathway that encodes experience  meaningfully as a parametric distribution in an embedding space. This idea makes experience replay feasible without  requiring a memory buffer to store task data.
The algorithm is able to accumulate knowledge  consistently to past learned knowledge as the parametric distribution in the embedding space is enforced to be shared across all tasks. Compared to model-based approaches that regularize the network to consolidate the important weights for past tasks, our approach is able to address catastrophic forgetting without limiting the learning capacity of the network. Future works for our approach may   extend to learning new tasks and/or classes with limited labeled data points and investigating how to select   the suitable network layer and its dimension as the embedding.

\section{Acknowledgment}
This material is based upon work supported by the United States Air Force and DARPA under Contract No. FA8750-18-C-0103. Any opinions, findings and conclusions or recommendations expressed in this material are those of the author(s) and do not necessarily reflect the views of the United States Air Force and DARPA. 
 

\bibliographystyle{named}
\bibliography{ijcai19}

\end{document}